\documentclass[runningheads]{llncs}
\usepackage{comment}
\usepackage{algorithm}
\usepackage{booktabs}
\usepackage{algorithmic}
\usepackage{pifont}
\usepackage{multicol}
\usepackage{multirow}
\usepackage{todonotes}
\usepackage{xcolor} 
\newcommand{\cmark}{\textbf{\ding{51}}} 
\newcommand{\xmark}{{\color{gray}\scriptsize -}} 
\newcommand{\omark}{{\color{gray}\scriptsize \ding{55}}}

\usepackage{amssymb}

\usepackage[T1]{fontenc}
\usepackage{amsmath}

\newcommand{\resource}{\textsc{JediKG}}
\newcommand{\extraction}{\textsc{JDeX}}
\newcommand{\dataset}{\textsc{JDSet}}

\usepackage{inconsolata}
\usepackage{tablefootnote}
\usepackage{graphicx}
\usepackage{hyperref}
\usepackage{color}

\usepackage{xspace}
\usetikzlibrary{positioning,calc}

\newcommand{\rot}[1]{\rotatebox{90}{#1}}

\begin{document}
\title{JediKG: Return of the Schema}
\subtitle{Building Complete Datasets for Machine Learning and Reasoning on Knowledge Graphs}
\titlerunning{JediKG: Building Complete Datasets from Knowledge Graphs}

%
%
\author{
    Ivan Diliso\inst{1}\orcidID{0009-0007-2942-202X} \and 
    Roberto Barile\inst{1}\orcidID{0009-0007-3058-8692} \and 
    Nicola Fanizzi\inst{1,2}\orcidID{0000-0001-5319-7933} \and
    Claudia d'Amato\inst{1,2}\orcidID{0000-0002-3385-987X}
}
\institute{
    Dipartimento di Informatica -- University of Bari Aldo Moro, Italy\\ 
    \email{i.diliso1@phd.uniba.it}, 
    \email{r.barile17@phd.uniba.it},\\
    \email{claudia.damato@uniba.it}, 
    \email{nicola.fanizzi@uniba.it}
    \and CILA -- University of Bari Aldo Moro, Italy
}

%
%

%
\maketitle

{\renewcommand\thefootnote{}%
\footnotetext{This preprint has not undergone peer review (when applicable) or any post-submission improvements or corrections. The Version of Record of this contribution is published in \emph{The Semantic Web -- ISWC 2026}, LNCS, vol.~17127, Springer, Cham, and is available online at \url{https://doi.org/10.1007/978-3-032-42026-8_21}}}          

\begin{abstract}
Benchmark datasets for Knowledge Graph Refinement typically contain only ground facts, discarding schema available in the source knowledge graph. 
This limits the evaluation of Neural-Symbolic methods that leverage expressive schemas.
We present \resource{} the first resource that systematically addresses these gaps through two components: \extraction{} a general configurable tool for extracting complete, portable, modular dataset from any \texttt{RDF}/\texttt{OWL} knowledge graph;
and \dataset{}, a curated suite of eleven datasets with varying expressivity, volume, and domain.
\extraction{} preserves full schema expressivity, employs standard \texttt{OWL} reasoning to detect and correct inconsistencies and materialize implicit knowledge, applies schema modularization, and provides utilities for integration with mainstream machine learning libraries such as PyTorch and PyKEEN.
We validated the completeness and portability of the datasets across several dimensions and preliminarily proved the utility of \resource{} in experimental workflows combining data-driven methods with schema-driven reasoning.


\end{abstract}


\section{Introduction}
\emph{Knowledge Graphs} (KGs)~\cite{hogan2021knowledge} are graph-based representations of relational \emph{data}.
They are often coupled with more general knowledge like a \emph{schema} (e.g., based on \emph{ontologies}) that makes the meaning of data explicit and enables automated reasoning services.
Although KGs have proved useful in different contexts~\cite{dong2019building,yago4}, they are inherently incomplete and often noisy, as their life-cycle primarily relies on semi-automated and distributed processes~\cite{hogan2021knowledge}.
This calls for effective solutions to \emph{Knowledge Graph Refinement} (KGR) problems.
In particular, the tasks that have been most extensively investigated concern relationships between individuals.
Prominent methods are grounded in data-driven \emph{Machine Learning} (ML) techniques, such as \emph{Knowledge Graph Embedding} (KGE) models~\cite{rossi2021knowledge}, which learn vector representations of KGs.

KGR methods, as grounded in ML, require curated \textit{benchmark datasets} for experimental evaluation.
Consequently, many such datasets have been released, typically derived from popular KGs, such as \emph{DB100K}~\cite{db100k} and \emph{YAGO3-10}~\cite{yago310}, derived from \emph{DBpedia} and \emph{YAGO3}, respectively.
As KGs may contain billions of assertions, datasets are usually created by sampling manageable fragments.
These benchmarks reflect current KGR solutions which usually focus solely on data and disregard schema.
This limitation hinders the involvement of emerging \emph{Neural-Symbolic} (NeSy) methods, which can integrate schema with purely data-driven approaches.
These approaches can leverage schema to enrich representations~\cite{jackermeier2024dual,zhapa2024lattice}, optimization constraints~\cite{jain2021improving,d2021injecting}, or explanations~\cite{barile_explanationlink_2024}.
The experimental evaluation of such methods calls for benchmarks that include both data and schema, as advocated in~\cite{damato_machinelearning_2023}.

A few datasets containing both schema and data have been released (see, e.g.,~\cite{db100kplus}).
In general, they have been obtained by starting from existing data fragments, such as DB100K, and attaching them to relevant schema fragments with the desired expressivity.
Regretfully, several challenges arise in this process, and there is still no consensus on how to address them.
As a result, a new dataset is derived for each proposed NeSy method, often from the same KGs, with an heuristic approach tailored to that specific method.
This severely hampers fair and reproducible comparisons among experimental results of different methods.

Primarily, existing proposals differ in terms of expressivity, which is generally limited.
Instead, we argue for preserving the full expressivity of the schema.
This holds even if NeSy KGR methods leverage only a few constructs, because discarded ones can still be useful for auxiliary evaluation and debugging phases.
Another difference concerns how to correct logical flaws, which is important when NeSy methods assume flawless knowledge, as such flaws lead to trivial reasoning~\cite{krotzsch2012description}.
Existing proposals also differ in how relevant schema fragments are selected, ways that are generally not grounded in proper \emph{modularization} methods.
This leads to adopt different schema fragments even for datasets from the same KG.
Another incoherent aspect is the question of whether, and to what extent, datasets should be enriched with knowledge derived from reasoning.
Moreover, these datasets adopt implementation-specific formats rather than standard KG serializations.
The partitioning of datasets into components also differs: many datasets are released as monoliths that mix different types of constructs.

To address these limitations, we propose \resource{}.
To the best of our knowledge, it is the first resource to rationalize the construction of KGR datasets.
\resource{} includes: \extraction{}, a \emph{general} and \emph{semi-automated} workflow for deriving schema-enhanced datasets on-demand; and \dataset{}, a curated suite of datasets obtained by applying \extraction{} to several KGs.
In particular, \extraction{} takes as input any KG in standard \emph{Semantic Web} languages.
Thus, it derives a dataset that is: \emph{complete}, covering both data and schema while preserving the original expressivity, and \emph{portable}, usable by both learning and reasoning software while satisfying requirements of learning settings, such as size and density.
The extracted dataset is then serialized in common Semantic Web formats.
Additionally, the datasets are released in multiple components concerning different constructs and purposes.
\extraction{} grounds on standards methods and tools for KGs ranging from merging and querying to reasoning.
It also includes a phase useful for and grounded in ML workflows, implemented in popular KGR libraries such as \texttt{PyKEEN}, by performing dataset splitting with prior checks on coverage and leakage and by providing utilities for data loading.
To foster automatic (re)use of \resource{}, all the (automated) steps of \extraction{} are implemented in an open-source Python library available on GitHub with \emph{interactive} and \emph{configurable} usage thoroughly documented.
\dataset{} consists of eleven datasets based on six different KGs with varying expressivity, size, and domain.
Particularly, it covers both general and more specialized domains.
An ongoing extension of the resource with new and different datasets and features is planned and documented.

We posit that \resource{} addresses the lack of consensus that characterize existing approaches.
The suite \dataset{} supports this claim as it encompasses different datasets generated via the same process.
We test for completeness and portability of our datasets by assessing and comparing to prior efforts their expressivity, size, density, and distribution of property types.
We also posit that, given completeness and portability, these datasets can lower the barrier for practitioners working at the intersection of data-driven learning and schema-driven reasoning.
This is proved through a proof-of-concept where we illustrate and discuss an experimental workflow where these phases interact and give hints that might not emerge through existing datasets.

In summary, the contribution of the paper is \resource{}, a resource that rationalizes the incoherent and heuristic practices in the construction of KGR datasets.
Particularly, it encompasses three levels:
\begin{itemize}
    \item \extraction{}: a general and semi-automated workflow to derive complete and portable datasets on-demand;
    \item a configurable and interactive \texttt{Python} implementation of \extraction{}, grounded on Semantic Web standards;
    \item \dataset{} a suite of eleven complete and portable datasets derived from six KGs; evaluated across different dimensions.
\end{itemize}

The rest of the paper is organized as follows.
In \S~\ref{sect:related_works}, we review related methods and resources.
In \S~\ref{sec:resource}, we detail \resource{}.
In \S~\ref{sec:poc}, we show a proof of concept of how \dataset{} datasets can facilitate ML and reasoning tasks.
In \S~\ref{sect:conclusions}, we make final remarks and outline future extensions.


\section{Related Works}\label{sect:related_works}

Several publicly available KGs have been employed to construct benchmark datasets for KGR methods.
General-domain resources include \emph{DBpedia}~\cite{dbpedia}, \emph{Wikidata} \emph{Freebase}~\cite{freebase}, \emph{YAGO3}~\cite{yago3}, \emph{YAGO4}~\cite{yago4}, \emph{WordNet}, and \emph{NELL}~\cite{nell} varying in schema expressiveness (if present) and construction methodology. 
In our work we focus on KGs where schema is available and implemented in \texttt{OWL2}, to support experimental evaluation of KGR methods leveraging schema and foster the adoption of Semantic Web technologies. 

Domain-specific KGs remain largely absent from standard KGR benchmarks.
Notable exception include application of KGR on the \emph{Gene Ontology KG} (GO-KG)~\cite{sousa2023explainable,goonto}, though the source expressive \texttt{OWL} ontology is flattened into propositional assertions during construction.
Other expressive domain-specific KGs, such as \emph{ArCo}~\cite{arco} targeting the cultural heritage domain, \emph{WHOW}~\cite{whow} focused on water consumption and quality, \emph{CSKG}~\cite{dessi2025cs} about computer science and \emph{ERA}\footnote{\url{https://www.era.europa.eu/domains/registers/era-knowlege-graph_en}} targeting railway implant specifications, have not been incorporated into standard KGR benchmarks. 
We argue domain-specific KGs as particularly relevant for assessing schema-aware methods given their rich and expressive schemas and, for the aforementioned examples, also impressive data volume.
Our proposal therefore includes multiple domain-specific KGs.

Benchmark datasets are typically extracted from large KGs by selecting dense fragments of property (relationship) assertions, discarding attributes and often the schema itself.
Widely used examples include \emph{FB15K}~\cite{bordes2013translating} and \emph{WN18}~\cite{bordes2013translating} and their leakage-free variants \emph{FB15k-237}~\cite{fb15k237} and \emph{WN18RR}~\cite{yago310}, as well as \emph{YAGO3-10}~\cite{yago310}, \emph{DB100K}~\cite{db100k}, \emph{DB50K}~\cite{db50k}, all of which discard schema information even if available in the source KG.

Several KGR works have proposed schema-enhanced datasets, but their coverage and construction pipelines remain limited and incoherent.
In~\cite{d2021injecting}, datasets derived from DBpedia or YAGO are enriched only with method-specific axioms, such as class disjointedness and equivalence and property characteristics, without clearly documenting if implicit knowledge is materialized.
In~\cite{yago39k}, \emph{YAGO3-39K} includes class assertions and taxonomy information, including materialized implicit assertions and subclass relations, but relies on a custom transitivity-based procedure rather than standard \texttt{OWL} reasoning.
Similarly, in~\cite{sematk}, YAGO3-39K and DB100K are extended with class assertions, taxonomy, and domain/range axioms, while retaining only typed entities.
It also introduces \emph{YAGO4-19K}, a YAGO4 subset obtained by retaining only property with a specified domain and range and having at least $10$ assertions.
Differently, \textit{YAGO4-20}~\cite{yago420} is another subset of YAGO4 extracted by retaining assertions involving individuals occurring in at least $20$ triples and by manually correcting a few logical flaws.
However, the dataset is not complete off-the-shelf: class assertions are present, but the full schema must be integrated manually.
Recently, in~\cite{db100kplus} \emph{DB100K+}, \emph{YAGO3-10+}, and \emph{NELL-995+} were proposed.
They enrich DB100K, YAGO3-10, and a fragment of a specific NELL version with class membership, class taxonomy, and property domain and range axioms.
Class information is further enriched through reasoning, although again via a non-standard custom implementation.
Overall, these datasets are difficult to compare and reuse because their construction is insufficiently standardized, reproducible, or documented, especially regarding axiom selection, correction, enrichment, and release format.
They are also commonly released as ID-mapped assertions, which are convenient for embedding models but poorly suited to Semantic Web tools.
Moreover, since these works address only restricted subsets of axioms, they produce datasets with limited expressiveness, even when the original KG relies on a richer ontology.

\section{The Proposed Resource \resource{}}
\label{sec:resource}

In \S~\ref{subsect:workflow}, we detail the \extraction{} the dataset derivation workflow; in \S~\ref{subsect:implementation}, we illustrate the implementation of \extraction{}; in \S~\ref{subsect:dataset}, we present the \dataset{} dataset suite.

\subsection{\extraction{}: Dataset Extraction Workflow}\label{subsect:workflow}

 \begin{figure}[t]
    \centering 
    \includegraphics[width=\linewidth]{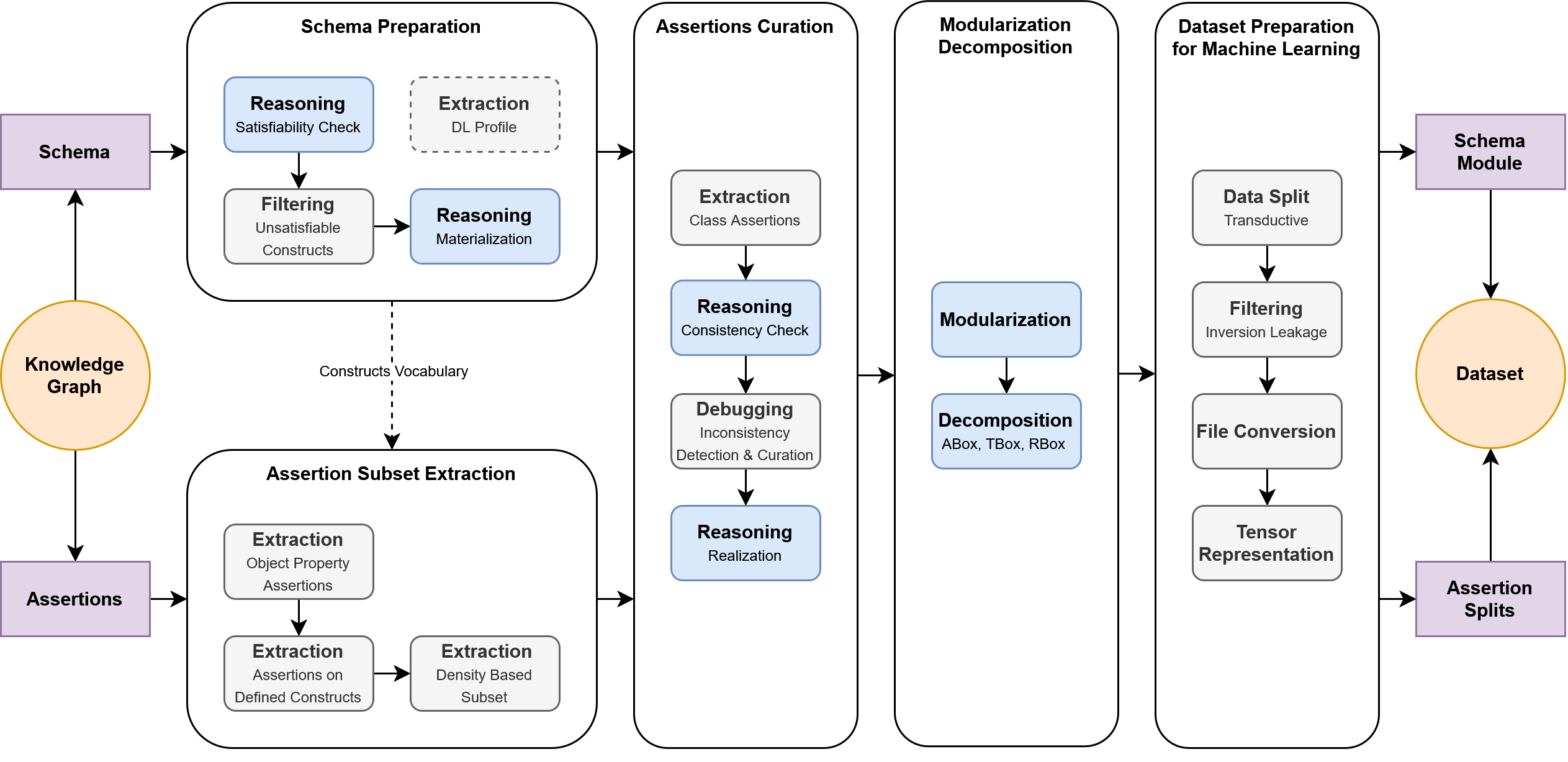}
    \caption{\extraction{} phases.}
    \label{fig:general_schema}
\end{figure}

\extraction{} is a general, interactive, configurable, and semi-automated workflow to derive, from large KGs, complete and portable datasets for KGR.
It applies to \texttt{OWL2}~DL KGs featuring: a schema, consisting of a \emph{TBox} for axioms about classes and a RBox for axioms about properties; a \emph{ABox} providing \emph{Ground Assertions} (or simply assertions) encompassing \emph{Class Assertions} (CAs), \emph{Object Property Assertions} (OPAs), and other kind of assertions regarding concrete datatypes and annotations.
It takes in input a KG in the form of a schema file and a assertions file.
The result is a dataset including both schema and data.
As shown in Fig.~\ref{fig:general_schema}, the process can be summarized into the following phases:
\begin{itemize}
    \item \textsf{Schema Preparation}: merge, correct, and enrich via reasoning the KG modular schema;
    \item \textsf{Assertions Subset Extraction}: select a more tractable subset of the assertions also w.r.t.\ the schema;
    \item \textsf{Assertions Curation}: correct and enrich the assertions subset via reasoning w.r.t.\ the pre-processed schema;
    \item \textsf{Schema Modularization}: select a schema fragment relevant to the curated assertions;
    \item \textsf{Dataset Decomposition}: decompose the resulting dataset into multiple modules;
    \item \textsf{Dataset Preparation for ML}: transform the resulting assertions for usage in ML workflows and convert the dataset components to serializations more friendly to widely used ML libraries.
\end{itemize}

\subsubsection{Schema Preparation}

In this phase, \extraction{} extracts a correct and enriched dataset schema from the full KG schema.
Since KG schemas are usually released in multiple modules, \extraction{} merges them into a single schema in order to leverage the the axioms of all modules in the subsequent phases.
Particularly, it collapses the import closure, i.e., it copies the axioms of all the imported schemas into one schema.
Then, it checks whether classes and properties are satisfiable (i.e., can have at least one instance in some possible model without causing a contradiction).
The schema may be corrected in order to apply reasoning services that require a correct schema in the subsequent phases. 
Furthermore, one may restrict the schema to a specific expressivity (e.g., to a specific DL with more tractable reasoning like $\mathcal{EL}$ or $\mathcal{ALC}$).
Finally, \extraction{} can compute the deductive closure of the schema w.r.t.\ a given set of axiom types, e.g., class subsumption or property characteristic axioms.

Concerning correction, it can be achieved by removing the unsatisfiable constructs.
An alternative is to manually inspect and modify the axioms of the unsatisfiable constructs via reasoning and justification. 
This would avoid the need to remove useful constructs.
Moreover, even if removing the unsatisfiable constructs leads to a satisfiable schema, the underlying modeling error may still persist and thus become masked rather than resolved.
\extraction{} defaults to removal in order to avoid additional assumptions that do not comply with the intended semantics of the schema.
Manual corrections may be applied when the practitioner using \extraction{} knows the intended semantics of the KG, or when constructs clearly violate the intended semantics (e.g., violations of DBpedia naming conventions causing reasoning issues).

\subsubsection{Assertions Subset Extraction}

This phase aims to extract a subset of assertions of tractable size (typically on the order of a few million assertions) with desirable properties such as density.
This is motivated by the scale of real-world KGs, which can range from millions to billions of assertions, making them impractical for experimental evaluation of KGR methods and tools.
\extraction{} first retains only OPAs, thus discarding data property assertions and annotation assertions, which are not typically considered KGR methods and tools.
Furthermore, only OPAs with properties declared in the schema are retained.
Hence, OPAs with properties found to be unsatisfiable, and thus removed from the schema, in the previous phase are excluded.

Next, \extraction{}, in order to make the dataset portable to ML, performs degree-based filtering that controls the size and density of the extracted subset.
Only OPAs whose individuals (both subject and object) appear in at least $k$ OPAs are retained.
The parameter $k$ thus controls both the size and the density of the resulting subset: higher values yield a smaller, denser core, while lower values preserve a larger but sparser subset.

\subsubsection{Assertions Curation}

Firstly, the CAs of the individuals remaining after filtering OPAs are selected and restricted to classes defined in the schema.
\extraction{} proceeds by removing ABox assertions that are closer to schema axioms than to factual assertions.
Such statements are frequently found in popular KGs and typically involve classes or properties as both arguments, effectively causing a class or property to also be treated as an individual.
We remove these assertions because they violate the separation between TBox and ABox, conflating the two.

Having obtained a satisfiable schema subset and a tractable assertion subset, \extraction{} merges them and corrects and enriches the resulting complete dataset.
Particularly, it performs consistency checking (whether a KG admits at least one model).
If the assertions are inconsistent w.r.t.\ the schema, a curation step is required before applying further reasoning services that assume a consistent input, such as CA realization.
The correction step is the only step that is not fully automated because debugging suggestions such as justifications~\cite{horridge2009explaining} (i.e., minimal sets of axioms that entail the inconsistency) are generally non unique and each contains several axioms: inconsistency resolution requires choosing, according to domain knowledge, which suggestion and which axioms therein should be leveraged for correction.
In principle, any semi-automated form of debugging, can be integrated in \extraction{}.
Differently, when fully automated correction is required, one must make additional assumptions, although it should be noted that the solution may be not optimal.
Once consistency has been restored, \extraction{} applies CA realization, which materializes the implicit CAs.

\subsubsection{Schema Modularization}

In this phase, \extraction{} performs \emph{ontology modularization}~\cite{grau2007logical}: it computes a schema module tailored to the assertions and preserving the same inferences that can be obtained by considering the whole schema.
Restricting the schema in this principled way also fosters the adoptions of the same schema fragment across different works: since the module is usually reasonable in size and contains all the schema axioms relevant to the assertions, practitioners do not need additional manual and ad-hoc filtering.
\extraction{} does not enforce minimality of the selected fragment, as the goal is only a more tractable fragment than the whole schema and avoiding minimality requirements also reduces the computational demands.
In principle, any other approach that extract a minimal module~\cite{grau2007just,romero2015ontology} can be integrated.

\subsubsection{Dataset Decomposition}
To further facilitate the use of the proposed datasets, \extraction{} decomposes the extracted complete dataset into multiple components.
This allows users to load only the components required in a given experimental setting.
In particular, each dataset is organized into TBox, RBox, and ABox components.
The TBox is further partitioned into the class taxonomy and the remaining non-taxonomic TBox axioms, whereas the ABox is partitioned into OPAs and CAs.

\subsubsection{Dataset Preparation for ML}

While the datasets serialized in \texttt{OWL2} remain fully and easily loadable in standard Semantic Web software, they are not yet portable to popular ML libraries such as \texttt{PyKEEN}~\cite{pykeen} or \texttt{PyTorch}~\cite{pytorch} requires additional steps and dealing with a different serialization.
For this purpose, several ML post processing steps are applied to ensure portability.

Firstly, \extraction{} splits the assertions into training, validation, and test sets according to given ratio.
It can also check a training coverage criterion typical of state-of-the-art KGR proposals.
Particularly, it ensures that all the assertions in the validation and the test set concern properties, classes, and individuals appearing in the training set.
This criterion ensures compatibility with both transductive and inductive KGR methods: while transductive methods require all evaluated individuals and properties to have been observed during training, inductive methods remain equally applicable when this condition holds.
Then, \extraction{} checks for inversion leakage~\cite{fb15k237}, which occurs when an assertion in the test or validation set is simply the inverse of one in the training set, which would allow the model to trivially infer test assertions without actually learning the underlying patterns, especially in methods where inverse assertions are used for data augmentation~\cite{kazemi2018simple}.
This filter removes test assertions that are inverses of training assertions.
Next, \extraction{} computes a tensor based representation of the dataset by assigning to each object property, individual, and class a univocal positive integer.
Finally, it converts some parts of the dataset into less expressive serializations, e.g., it converts OPAs in \texttt{tsv}, for higher portability to ML libraries.

\subsection{\extraction{} Implementation}
\label{subsect:implementation}

The proposed workflow is implemented as a (possibly) fully automated, configurable, and interactive CLI tool to execute all phases of \extraction{}.
Installation, configuration, and usage information are available in the official documentation and GitHub repository.

It is implemented in \texttt{Python}, particularly adopting widely used libraries such as \texttt{RDFLib}, \texttt{PyTorch}, and \texttt{PyKEEN}.
At its core, \extraction{} implements a wrapper for reasoning services, integrating multiple tools under a general \texttt{Python} API that simplifies the otherwise very heterogeneous interaction, including:  standalone reasoner \textsc{Konclude}~\cite{steigmiller2014konclude}, a very efficient parallel reasoner, and \textsc{Pellet}~\cite{sirin2007pellet}, which implements justification approaches; and the \textsc{ROBOT}~\cite{jackson2019robot}, a highly configurable \texttt{Java} tool for ontology management and reasoning that also integrates the \textsc{HermiT}~\cite{glimm2014hermit} and \textsc{ELK}~\cite{elkreason} reasoners.

As minimal input, \extraction{} requires two files: one containing the schema axioms and one containing the assertions.
Both files can be provided in any standard Semantic Web serialization, such as \texttt{N-Triples}, \texttt{OWL/XML}, \texttt{RDF/XML}, or \texttt{Turtle}.
The tool also takes in input a \texttt{JSON} file specifying the configuration parameters.
The output dataset is portable to Semantic Web as serialized in \texttt{RDF/XML} and divided in multiple files, the data files are also converted in \texttt{tsv}.

Particularly, the configuration allows to: 
\begin{itemize}
    \item selectively enable reasoning services and correction of logical flaws in \textsf{Schema Preparation} and \textsf{Assertions Curation}, allowing \extraction{} to be applied also to inconsistent datasets (but bypassing realization);
    \item configure the degree parameter $k$ for \textsf{Assertion Subset Extraction};
    \item specify, if reasoning is enabled, the preferred reasoner for each phase and the axiom types to consider in materialization during \textsf{Schema Preparation};
    \item enable \textsf{Schema Modularization} and \textsf{Dataset Decomposition}; 
    \item configure for \textsf{Dataset Preparation for ML} the ML split ratio and enable leakage and coverage checking, and their parameters;
\end{itemize}

\begin{algorithm}[t]
\caption{Signature Based Schema Modularization} \label{alg:modularization}
\begin{algorithmic}[1]
\REQUIRE Ontology $\mathcal{O}$, initial signature $\Sigma_0$
\ENSURE Extracted module $\mathcal{M} \subseteq \mathcal{O}$
\STATE $\Sigma \leftarrow \Sigma_0$; \quad $\mathcal{M} \leftarrow \emptyset$
\REPEAT
    \STATE $Q_1 \leftarrow \{\alpha \in \mathcal{O} \mid \operatorname{sig}(\alpha) \cap \Sigma \neq \emptyset\}$
    \STATE $\mathcal{M} \leftarrow \mathcal{M} \cup Q_1$
    \STATE $\Sigma \leftarrow \Sigma \cup \bigcup_{\alpha \in Q_1} \operatorname{sig}(\alpha)$
\UNTIL{$\Sigma$ does not change}
\RETURN $\mathcal{M}$
\end{algorithmic}
\end{algorithm}

The process begins by converting the input file to the standard \texttt{OWL/XML} serialization.
In the \textsf{Schema Preparation} phase the satisfiability check is performed via any of the aforementioned reasoning systems.
For the materialization of the deductive closure, \extraction{} resorts to \textsc{Hermit} or \textsc{ELK} via \textsc{ROBOT}. 
While this phase could be, in principle, be carried out with any reasoner, this is not directly implemented in the standalone reasoner themselves, but rather in tools that wrap and leverage them.
\textsc{ROBOT} was chosen for this purpose, due to its straightforward integration with \texttt{Python}.
Specifically, it allows materializing all the implicit axioms on class subsumption, classes equivalence, property equivalence, inverse properties, property hierarchies, property domains and ranges, and property characteristics (i.e., transitivity, functionality, symmetry, reflexivity).\footnote{\extraction{} defaults to include indirect inferences and exclude redundant or tautological axioms (e.g., a class subsumed by \texttt{owl:Thing})}

If the \textsf{Assertion Curation} phase is enabled, \textsc{Konclude} is used for consistency checking. 
Even if available in all the aforementioned tool, \textsc{Konclude} was preferred due to its efficiency.
For the same reasons, \extraction{} defaults to \textsc{Konclude}  for realization, though realization with \textsc{HermiT} or \textsc{ELK} is also supported. 
If inconsistencies are detected and fully automatic curation is required, \extraction{} currently supports a heuristic for automatic filtering based on justification: assuming the schema to be correct and assertions to be the cause of inconsistencies, \extraction{} filters out all assertions appearing in justification provided by \textsc{Pellet}, the only tool implementing justification.

Similarly, \extraction{} currently implements schema modularization via the seminal signature-based \textsc{PROMPTFactor}~\cite{promptfactor} algorithm, which we summarize in the following to make this paper more self-contained.
As formalized in Alg.~\ref{alg:modularization}, given an initial signature, it iteratively identifies all axioms from the input ontology $Q$ that overlap with the current signature, collects them into a sub-ontology $Q_1$, expands the signature with all symbols occurring in $Sig(Q_1)$, and repeats this process until a fixpoint is reached.
The initial signature is the set of all classes and properties used in the assertions (after realization).

Regarding portability to ML libraries, \extraction{} provides \texttt{PyTorch} data loaders with efficient handling of schema fragments as sparse tensors to minimize memory usage and ensure direct compatibility with \texttt{PyKEEN} training pipelines.
Additional utility functions enable efficient tensor-based traversal and manipulation.


\subsection{\dataset: Dataset Suite}\label{subsect:dataset}

In this section, we illustrate \dataset{}: the datasets resulting from the application of \extraction{}.
We specify the KGs adopted as sources and the configurations and then discuss the outcomes.

\subsubsection{Sources and Configuration}

We applied \extraction{} both to complete existing KG data fragments with schema and to derive brand-new complete datasets both in terms of schema and data.
When applied to existing data fragments, this produces a schema-enriched version of the dataset, facilitating the comparison of novel schema-aware methods with prior methods that were evaluated on the existing data fragments.
For these, the directly publicly available assertion subsets were used, thus skipping the subset extraction step of \extraction{}.

As source KGs, we selected six well-established, maintained, and large-scale KGs available in \texttt{RDF} and \texttt{OWL} covering both general-purpose and domain-specific knowledge.
Indeed, we include three general-purpose KGs: DBpedia~\cite{dbpedia}, YAGO~3~\cite{yago3} and YAGO~4~\cite{yago4}.\footnote{Although YAGO4.5~\cite{yago45} is the most recent version of YAGO, we excluded it because its schema is expressed in SHACL rather than OWL.}
These are very large in terms of data, but relatively limited in terms of schema expressivity (roughly $\mathcal{ALC}$)
We also include three domain-specific KGs that, to the best of our knowledge, have not never been used as sources for KGR datasets (neither their assertions nor their schema): ArCo~\cite{arco}, WHOW~\cite{whow}, and ERA.\footnote{\url{https://zenodo.org/records/15513855}}
These are recent, public, freely available, and reliable sources currently used and maintained by various trusted institutions.
Furthermore, they have large data volume and are coupled with very expressive schemas (also $\mathcal{SROIQ}$) including constructs that allow for interesting inferences otherwise not possible.
For instance, these include class disjointedness, and cardinality restrictions, from which true negative assertions, which are often missing in KGs, can be derived via reasoning, thus improving training of KGE models that currently rely on possibly false negatives~\cite{diliso25,damato_machinelearning_2023}.

We performed a manual correction phase to restore fully qualified URIs in (the data of) state-of-the-art datasets, where names were pre-preprocessed and mapped to integers, thus allowing to connect them to the schema.
We configured \extraction{} in order to derive from each KG multiple datasets with different values for the parameter $k$ in degree-based filtering (unless already imposed by the available data fragment) and for each one a version where reasoning and correction are enabled and one where they are disabled.
For the version with reasoning enabled, we chose all the axiom types except for class disjointedness.
Instead, the versions constructed without reasoning, the implicit knowledge is not materialized and the datasets may be inconsistent and have an unsatisfiable schema. 
We always enabled schema modularization and dataset decomposition.
For ML splitting, we adopted the pretty standard $80$/$10$/$10$ train/validation/test split ratio, ensuring proper coverage and no inversion leakage. 

All the files necessary to reproduce the construction of all datasets (initial assertions, schemas, and configuration files) are available in the repository.
\extraction{} can be run on commodity hardware, with reasoning phases presenting a negligible overhead ranging from seconds to minutes, as \textsc{Konclude} is used whenever possible.
Detailed execution times per phase and dataset are available in the documentation.

\begin{table}[t]
\footnotesize
\centering
\setlength{\tabcolsep}{4pt}

\caption{Schema axiom type coverage of different datasets. \omark\ denotes axioms available in the KG but not in the dataset.}
\label{tab:dataset_merged}

\begin{tabular}{@{}l@{\quad{}}ccccccc@{\qquad{}\qquad{}}ccccccc@{}}
\toprule
& \multicolumn{7}{c@{\qquad{}\qquad{}}}{\scriptsize\textbf{SOTA}}
& \multicolumn{7}{c}{\scriptsize\textbf{\dataset{}}} \\ 
\cmidrule(r{3.5em}){2-8}\cmidrule{9-15}

\textbf{OWL feature}
& \rot{DB100K}
& \rot{DB100K+}
& \rot{YAGO3-10}
& \rot{YAGO3-10+}
& \rot{YAGO3-39K}
& \rot{YAGO4-19}
& \rot{YAGO4-20}
& \rot{DB100K-C}
& \rot{YAGO3-10-C}
& \rot{YAGO3-39K-C}
& \rot{YAGO4-20-C}
& \rot{WHOW-5}
& \rot{ARCO-5}
& \rot{ERA-90}\\

\midrule

\tt ClassAssertion             
& \cmark & \cmark & \omark & \cmark & \cmark & \cmark & \cmark
& \cmark & \cmark & \cmark & \cmark &  \cmark & \cmark & \cmark \\

\tt SubClassOf                 
& \omark & \cmark & \omark & \cmark & \cmark & \cmark & \cmark
& \cmark & \cmark & \cmark & \cmark  & \cmark & \cmark  & \cmark \\

\tt EquivalentClasses          
& \xmark & \xmark & \xmark & \xmark & \xmark & \xmark & \xmark
& \xmark & \xmark & \xmark & \xmark  & \xmark & \cmark & \xmark \\

\tt DisjointClasses            
& \omark & \omark & \xmark & \xmark & \xmark & \omark & \omark
& \cmark & \xmark & \xmark & \cmark  & \cmark & \cmark & \cmark \\

\tt UnionOf                    
& \xmark & \xmark & \xmark & \xmark & \xmark & \omark & \omark
& \xmark & \xmark & \xmark & \cmark & \cmark & \cmark & \cmark \\

\tt IntersectionOf             
& \xmark & \xmark & \xmark & \xmark & \xmark & \xmark & \xmark
& \xmark & \xmark & \xmark & \xmark  & \xmark & \xmark & \xmark \\

\tt ComplementOf               
& \xmark & \xmark & \xmark & \xmark & \xmark & \xmark & \xmark
& \xmark & \xmark & \xmark & \xmark  & \xmark & \cmark & \xmark \\

\tt SomeValuesFrom    
& \xmark & \xmark & \xmark & \xmark & \xmark & \xmark & \xmark
& \xmark & \xmark & \xmark & \xmark  & \cmark & \cmark & \xmark \\

\tt AllValuesFrom      
& \xmark & \xmark & \xmark & \xmark & \xmark & \xmark & \xmark
& \xmark & \xmark & \xmark & \xmark  & \cmark & \cmark & \cmark \\

\tt CardinalityRestriction
& \xmark & \xmark & \xmark & \xmark & \xmark & \xmark & \xmark
& \xmark & \xmark & \xmark & \xmark  & \cmark & \cmark  & \cmark\\

\tt PropertyDomain            
& \omark & \cmark & \omark & \cmark & \omark & \cmark & \cmark
& \cmark & \cmark & \cmark & \cmark & \cmark & \cmark  & \cmark \\

\tt PropertyRange               
& \omark & \cmark & \omark & \cmark & \omark & \cmark & \cmark
& \cmark & \cmark & \cmark & \cmark & \cmark & \cmark  & \cmark\\

\tt SubProperty           
& \omark & \omark & \omark & \omark & \omark & \omark & \omark
& \cmark & \cmark & \cmark & \cmark  & \cmark & \cmark  & \cmark\\

\tt InverseProperty         
& \xmark & \xmark & \xmark & \xmark & \xmark & \omark & \omark
& \xmark & \xmark & \xmark & \cmark  & \cmark & \cmark  &  \cmark\\

\tt EquivalentProperty    
& \omark & \omark & \xmark & \xmark & \xmark & \xmark & \xmark
& \cmark & \xmark & \xmark & \xmark  & \cmark & \cmark  & \cmark\\

\tt PropertyCharacteristic   
& \omark & \omark & \omark & \omark & \omark & \omark & \omark
& \cmark & \cmark & \cmark & \cmark  & \cmark & \cmark & \cmark \\

\tt PropertyChain          
& \xmark & \xmark & \xmark & \xmark & \xmark & \xmark & \xmark
& \xmark & \xmark & \xmark & \xmark  & \cmark & \cmark  & \xmark\\

\bottomrule
\end{tabular}
\end{table}

\subsubsection{The Resulting Datasets}
The following presents the resulting datasets.
The brand-new datasets follow the naming convention \textit{\{KG\}--\{EX\}}, where \textit{KG} and \textit{EX} indicate the source KG and the $k$ value used in degree-based filtering, respectively.
In contrast, we refer to the enriched state-of-the-art datasets using the suffix \textit{-C}.
We release two complementary families of datasets, for a total of $11$ releases.
Specifically, $5$ datasets from general-purpose KGs, namely, \emph{DB-50K-C} from DB50K~\cite{db50k}, \emph{DB-100K-C} from DB100K~\cite{db100k}, \emph{YAGO3-10-C} from YAGO3-10~\cite{yago310}, \emph{YAGO4-20-C} from YAGO4-20~\cite{yago420}, and $6$ novel datasets from domain-specific KGs, namely, \emph{WHOW-5}, \emph{ARCO-5}, \emph{ARCO-10}, \emph{ARCO-20}, \emph{ERA-95}, \emph{ERA-90}.


Tab.~\ref{tab:dataset_merged} compares the expressivity of a selection of the datasets (one representative dataset per KG) in our proposed suite \dataset{} with widely used existing datasets.
As shown, recent efforts have aimed to release more comprehensive datasets; however, these typically focus on only basic constructs. 
In contrast, our datasets demonstrate that we preserve the full expressivity of the source KG after restricting it to tractable data and schema fragments.
To the best of our knowledge, these are the first ML-ready datasets based on KGs using highly expressive DLs, particularly $\mathcal{SROIQ}$ as in ArCo, and even including the often missing disjointedness axioms.
Another important aspect is that all the datasets have no logical flaws: we executed the consistency and satisfiability checks and corrected the resulting flaws.
This enables the usage of approaches that leverage reasoning services such as entailment that cannot be executed on inconsistent KGs. 
As recently advocated in~\cite{ERA_RailDataForum2025}, practical KG applications require not only data management, but also semantically rich resources that can, and must, support reasoning.
With \resource{}, we aim to foster the fulfillment of these requirements also when performing KGR in these critical domains.

Finally, the datasets are released in \texttt{OWL2~DL} serialized in \texttt{OWL/XML} and preserve the original namespaces and URIs.
The data part of the datasets is also serialized in \texttt{tsv} for easy usage with ML libraries such as \texttt{PyKEEN}.
This can be verified by downloading the datasets from our repository and parse them with standard software in both fields.

\begin{table}[t]
\footnotesize
\centering

\caption{Assertional-level statistics for newly generated datasets. Class assertions are reported as base / materialized counts.}
\label{tab:datasets_abox_stats}

\begin{tabular}{@{}lr@{\qquad{}}r@{\quad{}}r@{\qquad{}}rrr@{}}
\toprule

\multicolumn{1}{@{}l}{\multirow[c]{2}{*}{\bf Dataset}}
& \multirow[c]{2}{*}{\bf OPAs}
& \multicolumn{2}{c@{\qquad{}}}{\bf CAs}
& \multirow[c]{2}{*}{\bf Inds.}
& \multirow[c]{2}{*}{\bf Props.}
& \multirow[c]{2}{*}{\bf Concepts}

\\

&
& \bf \scriptsize pre-realiz.
& \bf \scriptsize post-realiz.
& 
&
&

\\

\midrule

DB50K-C &
28,516 & 12,413 & 171,348  & 22,263 & 278 & 372
\\

DB100K-C &
576,927 & 82,648 & 958,244 & 96,366 & 409 & 472
\\

YAGO3-39K-C &
369,987 & 441,990 & 968,271 & 37,711 & 35 & 46,892
\\
YAGO3-10-C &
1,080,369 & 1,309,963 & 3,124,625 & 123,038 & 35 & 94,726  
\\

YAGO4-20-C &
653,988 & 116,140 & 242,887  & 91,904 & 69 & 1,462
\\

WHOW-5 &
584,791 & 144,891 & 665,657 & 137,741 & 102 & 90
\\

ARCO-20 &
95,840 & 41,622 & 119,625 & 15,826 & 452 & 300
\\

ARCO-10 &
202,492 & 119,300 & 344,683 & 45,524 & 591 & 392
\\

ARCO-5 &
655,048 & 471,027 & 1,485,113  & 198,792 & 681 & 422
\\

ERA-95 &
465,451 & 18,046 & 86,267 & 18,046  & 80 & 41
\\

ERA-90 & 1,886,773 & 50,311 & 247,398 &  50,311 & 82 & 41
\\

\bottomrule
\end{tabular}
\end{table}

\begin{table}[t]
\footnotesize
\centering
\setlength{\tabcolsep}{10pt}

\caption{Relation cardinality statistics for newly generated datasets. The highest proportion for each dataset is highlighted in bold.}
\label{tab:datasets_cardinality_stats}

\begin{tabular}{@{}lrrrr@{}}
\toprule

\bf Dataset
& \bf 1-to-1
& \bf 1-to-N
& \bf N-to-1
& \bf N-to-N
\\

\midrule

DB50K-C &
\bfseries{0.545} & 0.029 & 0.418 & 0.007
\\

DB100K-C &
0.296 & 0.111 & \bfseries{0.362} & 0.232
\\

YAGO3-39K-C &
0.206 & 0.029 & \bfseries{0.412} & 0.353
\\

YAGO3-10-C &
0.059 & 0.147 & 0.294 & \bfseries{0.500}
\\

YAGO4-20-C &
0.221 & 0.147 & \bfseries{0.485} & 0.147
\\

WHOW-5 &
0.040 & 0.000 & \bfseries{0.920} & 0.040
\\

ARCO-20 &
0.075 & 0.113 & \bfseries{0.698} & 0.113
\\

ARCO-10 &
0.072 & 0.072 & \bfseries{0.703} & 0.153
\\

ARCO-5 &
0.144 & 0.072 & \bfseries{0.631} & 0.154
\\

ERA-95 &
0.000 & 0.025 & \bfseries{0.775} & 0.200
\\

ERA-90 &
0.000 & 0.024 & \bfseries{0.714} & 0.262
\\

\bottomrule
\end{tabular}
\end{table}

Tab.~\ref{tab:datasets_abox_stats} highlights the different structural profiles of the extracted datasets.
YAGO-derived datasets expose the richest class vocabularies, with YAGO3 in particular reaching tens of thousands of classes as also reflected in the large increase of class assertions after realization, where more than two million implicit memberships are materialized in YAGO3-10-C.
DBpedia and ArCo, instead, provide the largest property vocabularies, with ArCo reaching up to $681$ roles.
ERA shows a different profile: although it contains fewer classes and properties, it is highly connected at the ABox level, as reflected by the large number of OPAs obtained under a high-density setting.
These statistics show that the datasets are portable for usage in ML also in the sense that they provide a variety of profiles suitable to different target tasks: class-rich datasets are more suitable for evaluating type inference and NeSy methods taking into account class membership more extensively, datasets rich in terms of OPAs and properties better support relational prediction tasks, and denser datasets are preferable when connectivity is central to the evaluation.
In Tab.~\ref{tab:datasets_cardinality_stats} reports for each dataset the distribution of properties w.r.t.\ their cardinality, which is calculated with an heuristic~\cite{wang2014knowledge} implemented in \texttt{PyKEEN}.
N-to-1 relations dominate in most cases.
Notable exceptions are DB50K-C, where 1-to-1 relations are most frequent and YAGO3-10-C and WHOW-5, which exhibit a higher concentration of N-to-N and N-to-1 relations respectively, indicating a more complex relational structure. 
Taken together, structural proprieties, assertion density and individuals and class counts, highlight how each dataset poses its own learning challenges.

\section{Proof of Concept}
\label{sec:poc}

Here, we illustrate a proof-of-concept demonstrating that \resource{} enables workflows that combine purely data-driven ML and schema-driven reasoning.
Specifically, we show that they can be used to execute KGR methods based on purely data-driven KGE models and to perform an additional schema-driven validation phase where predictions are assessed w.r.t.\ the schema by checking if they are consistent with the dataset.
Such workflows are not possible with prior KGR datasets, where schema is missing or very limited.
Even if the schema is available, logical flaws are not always corrected, compromising schema-driven validation.

We therefore execute link prediction methods based on different KGE models on a selection of dataset in \dataset{}.
For this proof-of-concept, we adopted a (deliberately non-exhaustive) selection of four KGE models representative of different families: the geometrical models \textsc{TransE}~\cite{bordes2013translating} and \textsc{RotatE}~\cite{sun2019rotate}, the tensor-factorization model \textsc{ComplEx}~\cite{trouillon2016complex}, and the neural model \textsc{CompGCN}~\cite{vashishth2019composition}.
All models are trained with default training parameters and setup. 
The evaluation results are summarized in Tab.~\ref{tab:kgr_stats}, where we reported, for each dataset, only the best-performing model and the related metrics (well-established in KGR); a complete report is available in the online documentation.
This shows the application of an experimental workflow with a purely data-driven stance.

Next, we illustrate the schema-driven validation of link prediction outputs w.r.t.\ schema.
Particularly, we performed a test in a KG completion scenario, where plausible assertions are proposed (to be added in the KG), one at a time, according to a score computed by a KGE model.
For each proposal, we run a consistency check to verify whether the proposed assertion remain consistent with the prior knowledge.
This requires scoring all possible assertions, which scales combinatorially w.r.t.\ the number of individuals and properties.
For this reason, we executed this test with \textsc{TransE} on ARCO-10 as a computationally tractable yet sufficient proof-of-concept of enabling an hybrid workflow.
The results show that high model confidence, and high evaluation metrics, do not necessarily imply semantic validity: the KG remains consistent for the first 1425 added predictions, but becomes inconsistent after adding 1426 assertions.
Any further addition would be inconsistent given the monotonicity of reasoning in OWL.
This shows the execution of an hybrid workflow involving both purely data-driven and  facilitated by \dataset{}: we executed the reasoner seamlessly in a coherent workflow directly using the utilities implemented in \extraction{} and no additional manual steps.

\begin{table}[t]
\footnotesize
\centering
\setlength{\tabcolsep}{10pt}

\caption{Link prediction benchmarks on a sample of the proposed dataset. For each dataset only the best model (according to $H@10$) evaluation is shown}
\label{tab:kgr_stats}

\begin{tabular}{@{}lrrrrr@{}}
\toprule

\textbf{Dataset} 
& \textbf{Best Model}
& $\textit{MRR}$ 
& $H@1$
& $H@5$ 
& $H@10$ 

 \\

\midrule

DB100K-C &
\sc RotatE  & 0.371 & 0.282 & 0.468 & 0.541
\\

YAGO3-39K-C &
\sc RotatE   & 0.225 & 0.140  & 0.298  & 0.391
\\

YAGO4-20-C &
\sc RotatE  & 0.365 & 0.297  & 0.438  & 0.494
\\

WHOW-5 &
\sc RotatE  & 0.529 & 0.417  & 0.673  & 0.766
\\

ARCO-10 &
\sc RotatE & 0.759 & 0.680  & 0.845  & 0.867
\\

ERA-95 &
\sc CompGCN  & 	0.801	 & 0.714&	 0.909 &	0.960
\\

\bottomrule
\end{tabular}%
\end{table}

\section{Conclusion}\label{sect:conclusions}

In this work we introduced \resource{}, the first resource that systematically addresses the construction of datasets that combine both schema and data from large and expressive KGs.
Unlike existing benchmark datasets, which typically retain only data and are derived in method-specific ways, \resource{} delivers complete and portable datasets that are suitable for data-driven methods, schema-based reasoning, and even their combination.
The central component of this resource is \extraction{}, a general KG-independent workflow that refines expressive schemas, extracts controlled subsets of assertions, curates data, extracts schema modules tailored to the assertions, and performs final operations prior to ML workflows.
Being applicable to any \texttt{RDF}/\texttt{OWL} KG it is both reusable and extensible.
It has a \texttt{Python} implementation encompassing utilities for reasoning, ontology management, dataset splitting and tensor representations.
Building on this, we released \dataset{}, a curated suite of eleven datasets extracted from six different KGs.
We showed that the suite preserves the expressive power of the original schema, maintain controlled size, and exhibit a variety of useful properties.
Compared to state-of-the-art efforts for schema-enriched datasets, our releases retain schema more extensively, adopt standard Semantic Web syntaxes and can be processed directly by both reasoners and ML software.
Overall, \resource{} provides the groundwork for bridging the gap between schema-driven reasoning and data-driven learning.
As part of the ongoing development of the resource, we plan to extend \dataset{} with additional datasets from additional domain-specific KGs, incorporate support for datatype assertions, and explore alternative module extraction and curation strategies.
In the future, we plan to fully leverage this resource in order to integrate actual NeSy methods taking into account full expressivity.
We may also investigate additional dataset and schema representations that are fully compliant with parallel architectures, as conversion of \texttt{OWL} axioms to numerical representations is non-trivial.

\begin{credits}

\paragraph*{Resource Availability Statement:} 
 The complete archived resource is available on Zenodo under the MIT License at \href{https://doi.org/10.5281/zenodo.17817931}{https://doi.org/10.5281/zenodo.17817931}. The source code and resources are available through the GitHub repository \href{https://github.com/ara-t3/jedikg}{ara-t3/jedikg}. The official documentation is available at \href{https://ara-t3.github.io/jedikg/}{https://ara-t3.github.io/jedikg/}.

\subsubsection{\ackname} This work has been carried out within the framework of the PhD program \textit{Patti territoriali dell'alta formazione per le imprese} co-funded by AI$^2$ - Applicazioni di Ingegneria ed Informatica S.r.l.\ (CUP F61B23000370006) and of the MUR (Italian Ministry of University and Research) PhD program, D.M.\ n.~630:24.04.2024, co-funded by Sidea Group S.r.l.\ (CUP B91I24000240007).

\subsubsection{\discintname}
The authors declare to have no competing interests relevant to the content of this article.

\subsubsection{Declaration on Generative AI}
During the preparation of this work,  DeepL Write, Grammarly, and ChatGPT were used for grammar checking, rephrasing, and rewording.
After using these tools and services, the authors reviewed and edited the content as needed, taking full responsibility for the publication's content. 

\end{credits}
%
%
%
%

\bibliographystyle{splncs04}
\bibliography{bib}

\end{document}